# Efficacy of Modern Neuro-Evolutionary Strategies for Continuous Control Optimization

Paolo Pagliuca, Nicola Milano, Stefano Nolfi

Laboratory of Autonomous Robots and Artificial Life,
Institute of Cognitive Science and Technologies,
National Research Council, Roma, Italy

**Abstract**

*We analyze the efficacy of modern neuro-evolutionary strategies for continuous control optimization. Overall, the results collected on a wide variety of qualitatively different benchmark problems indicate that these methods are generally effective and scale well with respect to the number of parameters and the complexity of the problem. Moreover, they are relatively robust with respect to the setting of hyper-parameters. The comparison of the most promising methods indicates that the OpenAI-ES algorithm outperforms or equals the other algorithms on all considered problems. Moreover, we demonstrate how the reward functions optimized for reinforcement learning methods are not necessarily effective for evolutionary strategies and vice versa. This finding can lead to reconsideration of the relative efficacy of the two classes of algorithm since it implies that the comparisons performed to date are biased toward one or the other class.*

## 1    Introduction

Model-free machine learning methods made significant progress in the area of sequential decision making which involves deciding from experience the sequence of actions that can be performed in a certain environment to achieve a goal.

In the area of reinforcement learning, progress has been achieved primarily by combining classic algorithms with deep learning techniques for feature learning. Notable examples are agents trained to play Atari games on the basis of raw pixels input (Mnih et al., 2015) and simulated robots capable of performing locomotion and manipulation tasks (Schulman et al., 2015a and 2015b; Andrychowicz et al., 2019).

Recently, similar progress has been made in the area of evolutionary computation through neuro-evolutionary methods (Stanley, Clune, Lehman & Miikkulainen, 2019), also indicated as direct policy search methods (Schmidhuber & Zhao, 1999). In particular, in a recent paper Salimans et al. (2017) demonstrated how neural network controllers evolved through a specific natural evolutionary strategy achieve performance competitive with the reinforcement learning methods mentioned above on the MuJoCo locomotion problems (Todorov et al., 2012) and the Atari games from pixel inputs (Mnih et al., 2015).  In this work Salimans et al. (2017) also demonstrated for the first time that evolutionary strategies can be successfully applied to search spaces involving several hundred thousand parameters and can complete the evolutionary process in few minutes thanks to the their highly parallelizable nature.

However, the relation between the OpenAI-ES algorithm introduced by Salimans et al. (2017) and other related algorithms such as CMA-ES (Hansen & Ostermeier, 2001) and Natural Evolutionary Strategies (Wierstra et al., 2014) is still to be clarified. In particular, whether or not the former method is more effective than the other related methods, and/or whether the advantage of the method introduced by Salimans et al. (2017) comes from the usage of the virtual batch normalization technique (Salimans et al., 2016, 2017) that can be applied also to the other methods.

Such et al. (2017) compared the OpenAI-ES method with related algorithms. They used a classic evolutionary strategy (see the next section) and obtained performance similar to those reported by Salimans et al. (2017) on 13 selected Atari games, but lower performance on the MuJoCo humanoid locomotion problem. The classic method resulted less sample efficient than the natural evolutionary strategy used by Salimans et al. (2017) on this problem. Mania et al. (2018) demonstrated how a simplified evolutionary strategy is sufficient to solve the MuJoCo locomotion problems and outperform state-of-the-art policy gradient methods. Henderson et al. (2018) stressed the importance of considering the variability among replications and the impact of hyper-parameters to evaluate the efficacy of alternative methods.

Other works pointed out that the considered problems admit compact solutions. In particular, Rajeswaran et al. (2017) demonstrated how the MuJoCo locomotion problems can be solved with shallow networks. Such et al. (2017) demonstrated that some Atari games admit simple solutions, an issue highlighted also in other works (e.g., Wilson et al., 2018).

In this paper we compare systematically the performance of the evolutionary strategy proposed by Salimans et al. (2017) with other related methods in order to verify the relative efficacy of available algorithms on continuous optimization problems. To avoid biases caused by the usage of a specific class of problems, we extend the test with additional and qualitatively different problems (see below). Finally, we analyze the role of the reward function and critical hyper-parameters.

As we will see, the evolutionary strategy proposed by Salimans et al. (2017) outperforms or equals related approaches in all problems and is relatively robust with respect to the setting of hyper-parameters. The advantage of this method is not only due to the usage or virtual batch normalization that have been applied to all methods in our analysis. It can rather be ascribed to the efficacy of the Adam stochastic optimizer (Kingma et al., 2014) which avoids an uncontrolled growth of the size of the connection weights. Finally, we show how the contribution of virtual batch and weight decay normalization is minor in simple problems, but crucial in more complex ones.

The analysis of the role of the reward function indicates that functions optimized for reinforcement learning are not necessarily effective for evolutionary strategies and vice versa. Indeed, the performance of evolutionary strategies can improve dramatically with the usage of suitable reward functions. This finding should lead to a reconsideration of the relative efficacy of the two classes of algorithm since it implies that the comparisons performed to date are biased toward one or the other class.

## 2    Methods

In this section we briefly review the algorithms used in our experiments.

Evolution Strategies (ES), introduced by Rechenberg and Schwefel (Rechenberg and Eigen, 1973, Schwefel, 1977), were designed to cope with high-dimensional continuous-value domains and have remained an active field of research since then. They operate on a population of individuals (in our case, a population of vectors encoding the parameters of corresponding neural network policies). Variations are introduced in the policy parameters during the generation of new individuals. At every iteration ('generation'), the performance of the individuals with respect to an objective function ('fitness' or cumulative reward) is evaluated, the best individuals are kept, and the remaining individuals are discarded. Survivors then procreate by creating copies of themselves with mutations (i.e. variations of parameters). The process is repeated for several generations. We refer to these methods as classic evolutionary strategies.

This algorithm framework has been extended over the years to include the representation of correlated mutations through the use of a full covariance matrix. This led to the development of the **CMA-ES** (Hansen & Ostermeier, 2001) algorithm that captures interrelated dependencies by exploiting covariance while 'mutating' individuals. The algorithm estimates the covariance matrix incrementally across generations, thus extracting information about the correlation between consecutive updates. The matrix is then used to generate a parametrized search distribution.

Natural Evolutionary Strategies (Wierstra et al., 2014) are a variant of the CMA-ES that also rely on a parametrized search distribution based on a covariance matrix, and use the fitness of the population to estimates the local variation of the fitness function, i.e. the search gradient on the parameters toward higher expected fitness. Then they perform a gradient ascent step along the natural gradient by using a second-order method that renormalizes the update with respect to uncertainty. Natural evolutionary strategies came in two varieties: Exponential Natural Evolutionary Strategy (**xNES,** Wierstra et al., 2014) and Separable Natural Evolutionary Strategy (**sNES,** Wierstra et al., 2014). The latter is a simplified version that estimates the covariance of the diagonal instead of the full matrix and, consequently, scales-up to larger search spaces.

The **OpenAI-ES** method proposed by Salimans et al. (2017) is a form of natural evolutionary strategy that estimates the gradient of the expected fitness. Unlike the xNES and sNES, it performs mutations by using a simple isotropic Gaussian distribution with fixed variance. It uses the fitness of the population to estimate the gradient and updates the center of the distribution of the population through the Adam stochastic gradient optimizer (Kingma et al., 2014).

The OpenAI-ES method (Salimans et al., 2017) also relies on virtual batch normalization and weight decay. The former permits to normalize the distribution of the activation of the sensors. It is a variation of the batch normalization method commonly used in supervised learning adapted to problems in which the stimuli experienced by the network are not fixed (see also Salimans et al., 2016). This is the case of embodied agents in which the stimuli that are experienced depend on the actions executed by the agents previously. The problem is solved by calculating the average and the variance of the activation of the sensors incrementally on the basis of the distribution of the activation of the sensors of agents of successive generations. This technique is particularly useful in problems in which the range of activation of the sensors vary widely during the course of evolutionary process. Weight decay is a regularization technique, also commonly used in supervised learning, that penalizes the absolute value of weights to favor the development of simpler solutions that are less prone to overfitting. More specifically the OpenAI-ES method relies on an L1 weight decay and reduces the absolute value of the weights of 5‰ every generation.

We refer to these extended methods as modern evolutionary strategies. More specifically, we use this term to indicate algorithms computing the interrelated dependencies among variations of individuals or using a form of finite difference method to estimate the local gradient of the fitness function.

To analyze the efficacy of different reward functions for evolutionary strategies and reinforcement learning algorithms we used the Proximal Policy Optimization algorithm (**PPO**, Schulman et al. 2017). PPO is a state-of-the art policy gradient method (Peters and Schaal, 2008), a class of algorithms particularly suitable for the optimization of neural network policies applied to continuous control problems. PPO operates on a single individual policy and introduces variations by using stochastic actions. As the related TRPO algorithm (Schulman et al. 2015), PPO achieves learning stability by ensuring that the deviation from the previous policy is sufficiently small during parameter's update.

The source code that can be used to replicate the experiments described in Section 4 is available from https://github.com/PaoloP84/EfficacyModernES. The source code that can be used to replicate the experiment described in Section 5 and 6 is available from https://github.com/snolfi/evorobotpy/. The implementation of the algorithms has been based on the free software made available from the authors (i.e. http://pybrain.org/ for xNES and sNES, http://cma.gforge.inria.fr/ for CMA-ES, https://github.com/openai/evolution-strategies-starter for OpenAI-ES, https://github.com/openai/baselines for PPO). In the case of baseline implementation of PPO we did not need to introduce any change. The source code of the OpenAI-ES algorithm is designed to run on an amazon cluster environment. We adapted it to run on a standard linux machine. In the case of source code of CMA-ES, xNES and sNES, we created a standalone version of the original code integrated with a neural network simulator and with the OpenAI Gym environment (https://gym.openai.com/)

### 3    Problems

In this section we review the problems used in our experiments.

The first five considered problems are the MuJoCo locomotion problems (Todorov, Erez and Tassa, 2012) available in the Open-AI Gym environment (Brockman et al., 2016), which are commonly used as a benchmark for continuous control optimization. In particular, we used the **Swimmer** (Purcell, 1977; Coulom, 2002), the **Hopper** (Murthy and Raibert 1984; Erez, Tassa and Todorov, 2012), the **Halfcheetah** (Wawrzynski, 2007), the **Walker2D** (Raibert and Hodgins, 1991; Erez, Tassa and Todorov, 2012) and the **Humanoid** (Erez, Tassa and Todorov, 2012) problems. These tasks consist of controlling articulated robots in simulation for the ability to locomote as fast as possible by swimming in a viscous fluid (Swimmer), hopping (Hopper) and walking (Halfcheetah, Walker2D, and Humanoid).

The Swimmer, Hopper, Halfcheetah, Walker2D and Humanoid are provided with 2, 3, 6, 6, and 17 actuated joints, respectively. The observation state varies from 3 to 376 states. The observation includes the position and orientation of the robot, the angular position and velocity of the joints, and (in the case of the Humanoid) the actuators and external forces acting on the joints. The action state includes N values encoding the torques applied to the N corresponding joints. The initial posture of the robot varies randomly within limits. The evaluation episodes are terminated after 1000 steps or, prematurely, when the torso of the robots falls below a given threshold in the case of the Hopper, Walker2D and Humanoid. The agents are rewarded proportionally to their speed toward a target destination. However, they also receive additional rewards and punishments to facilitate the development of the required behaviors. More precisely, the agents are rewarded with a constant value for every step spent without falling (in the case of the Hopper, Walker2D, and Humanoid), and are punished with a quantity proportional to the square of the torques used to control the joints. In the case of the Humanoid, the robot is also punished with a quantity proportional to the square of the external forces acting on the joints. For other details see the references above.

The sixth considered problem is the **Long double-pole balancing** problem (Pagliuca, Milano and Nolfi, 2018; https://github.com/snolfi/longdpole) which consists in controlling a cart with two poles, attached with passive hinge joints on the top side of the cart, for the ability to keep the poles balanced. The cart has a mass of 1 Kg. The long pole and the short pole have a mass of 0.5 and 0.25 Kg and a length of 1.0 and 0.5 m, respectively. The cart is provided with three sensors encoding the current position of the cart on the track (x), and the current angle of the two poles ($\theta_1$ and $\theta_2$). The activation state of the motor neuron is normalized in the range [-10.0, 10.0] N and is used to set the force applied to the cart. The initial state of the cart is selected randomly at the beginning of every evaluation episode within the following intervals: $[-1.944 < x < 1.944, -1.215 < \dot{x} < 1.215, -0.0472 < \theta_1 < 0.0472, -0.135088 < \dot{\theta}_1 < 0.135088, -0.10472 < \theta_2 < 0.10472, -0.135088 < \dot{\theta}_2 < 0.135088]$. Episodes terminate after 1000 steps or when the angular position of one of the two poles exceeded the range $[-\frac{\pi}{5}, \frac{\pi}{5}]$ rad or when the position of the cart exceed the range [-2.4, 2.4] m. It is a much harder version than the classic non-markovian double pole balancing problem (Wieland, 1991) in which: (i) the length and the mass of the second pole is set to $\frac{1}{2}$ of that of the first pole (instead of $\frac{1}{10}$), and (ii) the agent should balance the poles from highly variable initial states. The reward consists of a constant value gained until the agent manages to avoid the termination conditions. The state of the sensors, the activation of the neural network, the force applied to the cart, and the position and velocity of the cart and of the poles are updated every 0.02 s. Unlike the MuJoCo locomotion tasks, this problem necessarily requires memory. For more details, see Pagliuca, Milano and Nolfi (2018) and Pagliuca and Nolfi (2019).

The seventh problem is the **Swarm foraging** problem (Pagliuca and Nolfi, 2019) in which a group of 10 simulated MarXbots (Bonani et al., 2010) should explore their environment so to maximize the number of food elements collected and transported to a nest. The robots are located in a flat square arena of 5 x 5 m, surrounded by walls, which contains a nest, i.e. a circular area with diameter of 0.8 m painted in gray. The robots, which have a circular shape and a diameter of 0.34 m, are provided with two motors controlling the desired speed of the two corresponding wheels, a ring of LEDs located around the robot body that can be turned on or off and can assume different colors, an omnidirectional camera, 36 infrared sensors located around the robot body that can detect the presence of nearby obstacles, and 8 ground infrared sensors that can be used to detect the color of the ground. Four hundred elements of invisible food are located inside 400 corresponding 0.5 x 0.5 m non-overlapping portions of the environment. The robots have an energy level that is replenished inside the nest and decreases linearly over time outside the nest. More specifically, the energy level is reset to 1.0 inside the nest and decreased of 0.01 every 100 ms spent outside the nest. To collect food, the energy of the robot should be greater than 0.0. Food elements are automatically collected and released when the robot enters in a portion of the environment containing a food element and in the nest, respectively. Effective solutions of this problem include robots capable of generating and exploiting specific spatial configurations, communicating with the other robots by turning on and off colored LEDs and by reacting to perceived colors, and assuming complementary different roles. The observation state includes 19 values encoding the state of the infrared sensors, of the ground sensors, of portions of their visual field, and of the battery of the robot. The action state includes four values encoding the desired speeds of the left and right robot's wheels, and the state of the blue and red LEDs located on the frontal and rear side of the robot. For other details, see Pagliuca and Nolfi (2019).

Finally, we used the Pybullet (Coumans and Bai, 2016) locomotion problems that constitute a free alternative to the MuJoCo environments re-tuned to produce more realistic behaviors. More specifically, we use the **HopperBullet**, **HalfcheetahBullet**, **Walker2DBullet, AntBullet**, and **HumanoidBullet** problems. Like in the MuJoCo versions, the robot has 3, 6, 6, 8, 17 actuated joints, respectively. The observation state varies from 15 to 44 states. The observations include the position and orientation of the robot, the angular position and velocity of the joints, and the state of contact sensors located on feet. The action state includes N values encoding the torques applied to the N corresponding joints. The initial posture of the robot varies randomly within limits. The evaluation episodes are terminated after 1000 steps or, prematurely, when the position of the torso of the robots falls below a given threshold in the case of the HopperBullet, Walker2DBullet, AntBullet and HumanoidBullet. The reward functions included in Pybullet and the varied reward functions optimized for evolutionary methods are tried are described in Section 5.

## 4 Comparative performance of evolutionary strategies

In this section we analyze the efficacy of the CMA-ES, xNES, sNES and OpenAI-ES methods on the MuJoCo locomotion problems, Long double-pole balancing, and Swarm foraging problems.

For the MuJoCo problems, we used the same parameters reported in Salimans et al. (2017). The neural network policy is a feed-forward network with 2 internal layers including 256 neurons in the case of the Humanoid and with a single internal layer including 50 neurons in the case of the other problems. The internal neurons use the hyperbolic tangent (tanh) function. The output neurons are linear. The state of the sensors is normalized through the virtual batch method described by Salimans et al. (2016, 2017). In the case of the Swimmer and the Hopper, the actions are discretized into 10 bins. Actions are perturbed with the addition of Gaussian noise with standard deviation 0.01. The evolutionary process lasted $2.5 * 10^8$ steps in the case of the Humanoid and $5 * 10^7$ steps in the case of the other problems. Agents are evaluated for 1 episode lasting up to 1000 steps.

In the case of the Long double-pole and Swarm foraging problem we used fully recurrent neural networks with 10 internal neurons. The internal and output neurons use the tanh function. The evolutionary process is continued for $1 * 10^{10}$ steps and for $1.5 * 10^6$ steps in the case of the Long double-pole and Swarm foraging problem, respectively. Agents are evaluated for 50 and 6 episodes in the case of the Long double-pole and Swarm foraging problems, respectively.

The number of connection weights and biases, which are encoded in genotypes and evolved, varies from a minimum of 1206, in the case of Halfcheetah and Walker2D, to a maximum of 166,673, in the case of the Humanoid. The population size was set to 500 in the case of the Humanoid and to 40 in the case of the other problems. For an analysis of the role of this and other hyper-parameters see Section 5. A detailed description of all parameters is included in the Appendix.

Table 1. Reward of the best evolved agents obtained with the different algorithms. Data indicate the average cumulative reward collected during an episode by the agents of each replication during a post-evaluaton test. The MuJoCo experiments have been replicated 20 times (10 in the case of the Humanoid). The long double-pole and the swarm foraging experiments have been replicated 50 and 30 times, respectively.

|  | CMA-ES | xNES | sNES | OpenAI-ES |
|---|---|---|---|---|
| Swimmer | **334.41 ± 72.50** | **364.73 ± 19.96** | **357.72 ± 45.41** | **347.84 ± 46.97** |
| Hopper | **3085.78 ± 661.80** | **3194.59 ± 667.53** | **3171.33 ± 578.23** | **3284.11 ± 764.20** |
| Halfcheetah | 3006.40 ± 1165.19 | 2602.42 ± 918.74 | 2890.44 ± 929.49 | **6556.97 ± 942.38** |
| Walker2D | 3215.97 ± 601.19 | 3255.69 ± 650.88 | 2986.65 ± 819.66 | **5086.67 ± 790.68** |
| Humanoid | n/a | n/a | 3993.92 ± 992.97 | **5661.30 ± 1398.72** |
| Long double-pole | 0.48 ± 0.084 | 0.622 ± 0.037 | 0.389 ± 0.087 | **0.752 ± 0.026** |
| Swarm foraging | 185.72 ± 22.80 | 204.51 ± 18.28 | 212.64 ± 27.95 | **282.97 ± 22.7** |

The analysis of the rewards obtained by the best evolved agents (Table 1) indicates that the OpenAI-ES algorithm outperforms the CMA-ES, sNES and xNES algorithms in the Halfcheetah, Walker2D, Humanoid, Long double-pole and Swarm foraging problems (Mann-Whitney U test with Bonferroni correction, p-value < 0.05) and achieves equally good performance on the Swimmer and Hopper problems (Mann-Whitney U test with Bonferroni correction, p-value > 0.05). In the case of the Humanoid problem, the high number of parameters makes the computation of the covariance matrix required for the CMA-ES and xNES unfeasible. Consequently, in the case of the Humanoid, we report the experiments carried with the sNES and OpenAI-ES methods only. Numbers in bold indicate the conditions leading to the best results. The variations of performance during the evolutionary process are displayed on Figure 1.

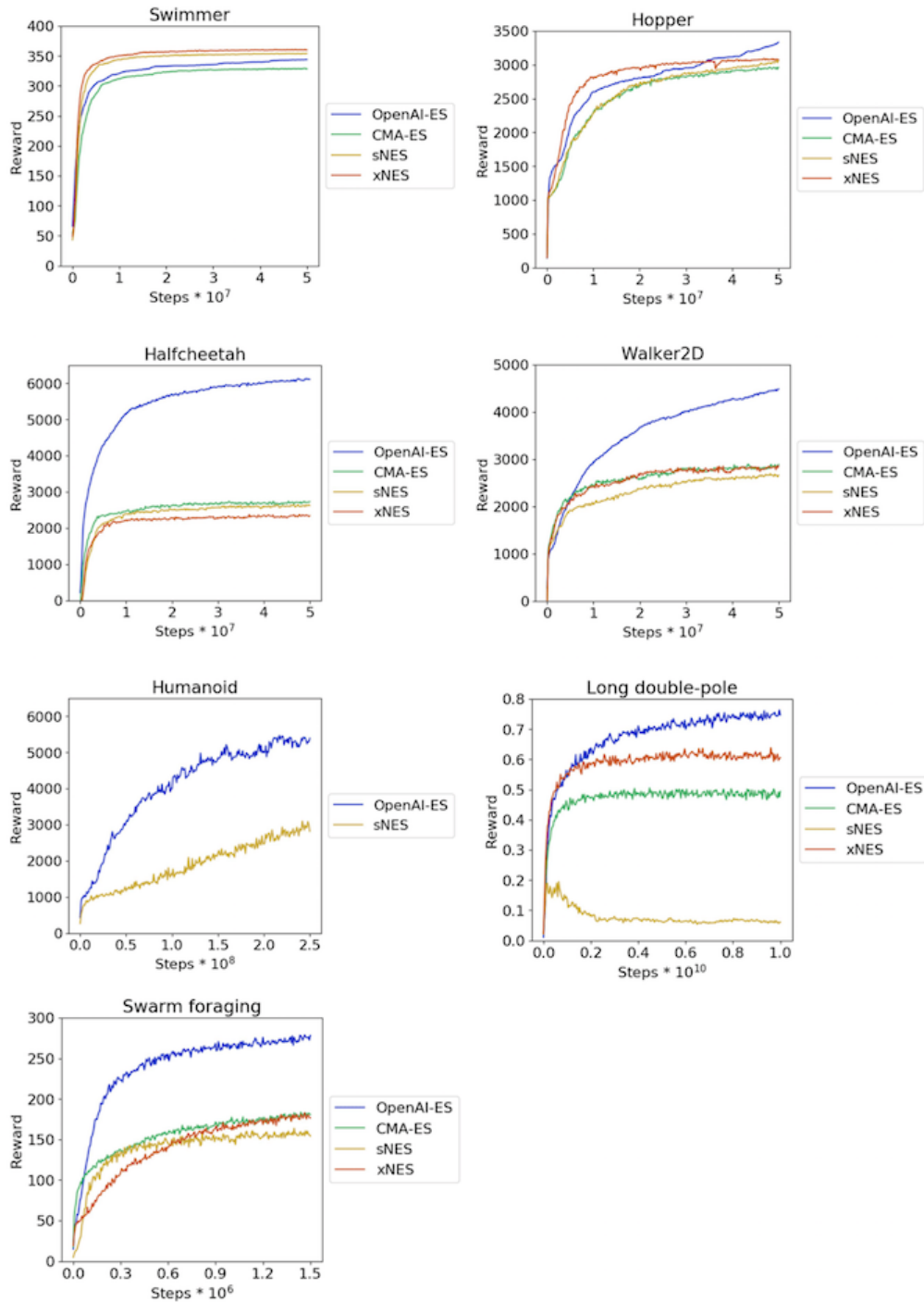

Figure 1. Reward obtained on the MuJoCo locomotion problems, on the long double-pole problem, and on the collective foraging problem during the course of the training process. Results obtained with the CMA-ES, sNES, xNES, and OpenAI-ES methods. Data indicate the average cumulative reward collected during an episode by the agents of each replication during a post-evaluaton test. Mean and 90% bootstrapped confidence intervals of the mean (shadow area).

Table 2. Average absolute size of connection weights of the best evolved individuals. Data averaged over multiple replications. Weight decay is used only in the experiment performed with the OpenAI-ES method with the robot locomotors problems.

|                  | CMA-ES          | xNES              | sNES              | OpenAI-ES        |
|------------------|-----------------|-------------------|-------------------|------------------|
| Swimmer          | 5.09 ± 3.885    | 13.906 ± 10.682   | 13.187 ± 10.11    | 0.17 ± 0.148     |
| Hopper           | 6.443 ± 4.96    | 17.356 ± 13.209   | 16.234 ± 12.459   | 0.169 ± 0.149    |
| Halfcheetah      | 4.333 ± 3.408   | 13.549 ± 10.59    | 11.068 ± 8.703    | 0.185 ± 0.367    |
| Walker2D         | 5.632 ± 4.417   | 17.899 ± 14.081   | 14.513 ± 11.473   | 0.184 ± 0.237    |
| Humanoid         | n/a             | n/a               | 6.531 ± 5.154     | 0.11 ± 1.524     |
| Long double-pole | 3.283 ± 2.727   | 51.698 ± 40.267   | 20.188 ± 16.788   | 0.645 ± 0.619    |
| Swarm foraging   | 3.987 ± 3.081   | 35.249 ± 28.307   | 105.099 ± 521.42  | 0.557 ± 0.454    |

As shown in Table 2, the absolute size of the connection weights grows significantly during the course of the evolutionary process in the case of the CMA-ES, xNES, and sNES methods, while it remains much smaller in the case of the OpenAI-ES method. This result is obtained independently of the usage of weight decay, as demonstrated by comparing the size of the solutions obtained with weight decay (Swimmer, Hopper, Halfcheetah and Walker2D) and without weight decay (Long double-pole and Swarm foraging).

Keeping the weight size small is important to preserve gradient information and reduce overfitting. Consequently, this property of the OpenAI-ES method can explain, at least in part, why it outperforms or equals alternative methods.

## 5   Sensitivity to the reward function

In this section we report the results obtained with the OpenAI-ES method on the Pybullet locomotors problem and the results of the analysis conducted by varying the characteristic of the reward functions.

The reward functions implemented in Pybullet (Coumans and Bai 2016) have been designed for reinforcement learning and are calculated by summing six components: (1) a progress component corresponding to the speed toward the destination, (2) a bonus for staying upright, (3) an electricity cost that corresponds to the average of the dot product of the action vector and of the joint speed vector, (4) a stall cost corresponding to the average of the squared action vector, (5) a cost proportional to the number of joints that reached the corresponding joint limits, and (6) a cost of -1.0 for falling to stay upright. The bonus is set to 2.0 in the case of the humanoid and to 1.0 in the case of the other problems. The electricity cost, stall cost, and joint at limit cost are weighted by -8.5, -0.425 and -0.1 in the case of the humanoid, and for -2.0, -0.1, and -0.1 in the case of the other problems. In principle the first component could be sufficient to learn a walking behavior. However, as we will see, additional components can be necessary.

As in the case of the MuJoCo locomotion problems, we used a feed-forward network with two internal layers including 256 neurons in the case of the HumanoidBullet and with a single internal layer including 50 neurons in the case of the other problems. The internal and output neurons use a tanh and a linear activation function, respectively. The state of the sensors is normalized through the virtual batch method described by Salimans et al. (2017). The evolutionary or learning process is continued for $5 * 10^7$ steps.

In the case of the experiments performed with the OpenAI-ES, the agents are evaluated for three episodes in the case of the HumanoidBullet problem and for one episode in the case of the other problems. As in the case of the MuJoCo locomotors experiments reported in the previous sections, the policy is deterministic but action states are perturbed slightly with the addition of Gaussian noise with 0.0 mean and 0.01 standard deviation.

In the case of the experiments performed with the PPO we used the default parameters included in the baseline implementation. The connection weights are updated every 2048 steps. The number of episodes depends on the number of restarts necessary to cover 2048 steps. The policy is stochastic and the diagonal distribution of Gaussian noise applied to actions is adapted together with the connection weights of the policy network.

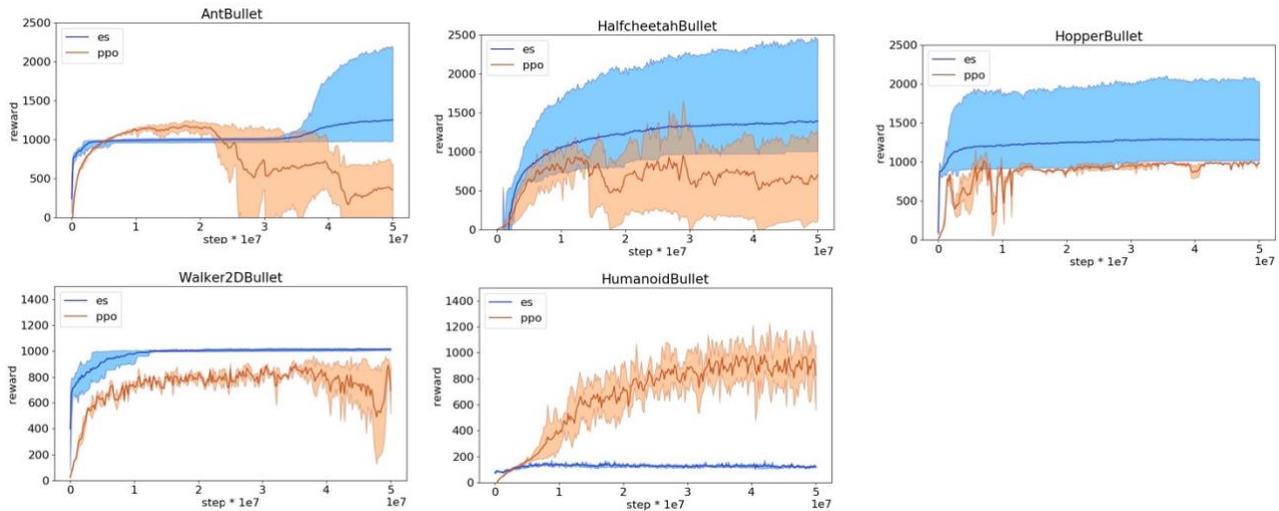

Figure 2. Reward obtained on the Pybullet problems during the training process with the reward functions optimized for reinforcement learning. Data obtained with the OpenAI-ES (es) and the PPO algorithms. Mean and 90% bootstrapped confidence intervals of the mean (shadow area) across 10 replications per experiment.

The obtained results indicate that the OpenAI-ES evolutionary strategy outperforms the PPO reinforcement learning algorithm in the case of the HopperBullet, HalfCheetahBullet, AntBullet and Walker2DBullet. Instead, PPO outperforms OpenAI-ES in the case of the HumanoidBullet (Figure 2). The inspection of the evolved behavior, however, indicates that the agents evolved with the OpenAI-ES algorithm are rather poor with respect to the ability to walk toward the target. This is illustrated by data reported in Figure 3 that shows the performance, i.e. the average distance in m travelled toward the target destination. The robots trained with the PPO display a much better ability to walk toward the destination.

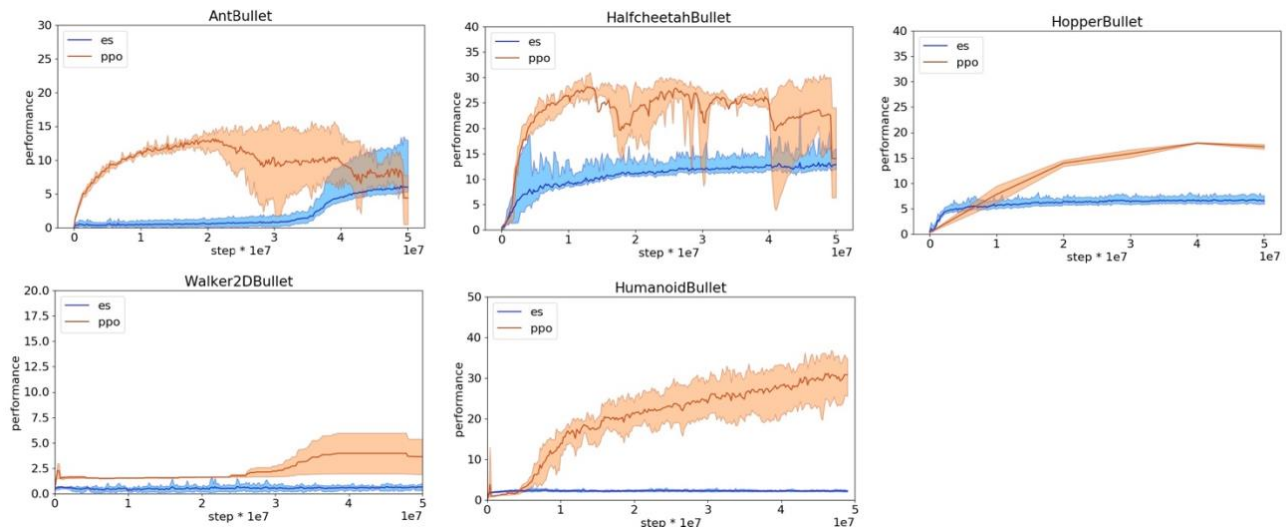

Figure 3. Performance obtained on the Pybullet problems with reward functions optimized for reinforcement learning method. Data obtained with OpenAI-ES (es) and PPO algorithms. Performance refers to the distance travelled toward the target destination during an episode in meters. Mean and 90% bootstrapped confidence intervals of the mean (shadow area) across 10 experiments per run.

The high reward achieved by the agents evolved with the OpenAI-ES method is due to their ability to maximize the steps in which they stay upright and to minimize the other costs. In the case of the HumanoidBullet, this is realized by moving on the place to postpone the falling as much as possible. In the case of the other problems, this is realized by assuming a posture from which the agent can remain still without moving, in most of the replications. In other words, in the case of the OpenAI-ES method, the usage of the bonus for staying upright and the other costs does not facilitate the development of an ability to walk effectively but rather drives the evolutionary process toward solutions that optimize the additional reward components without optimizing the first component that rates the agents for their ability to walk toward the destination.

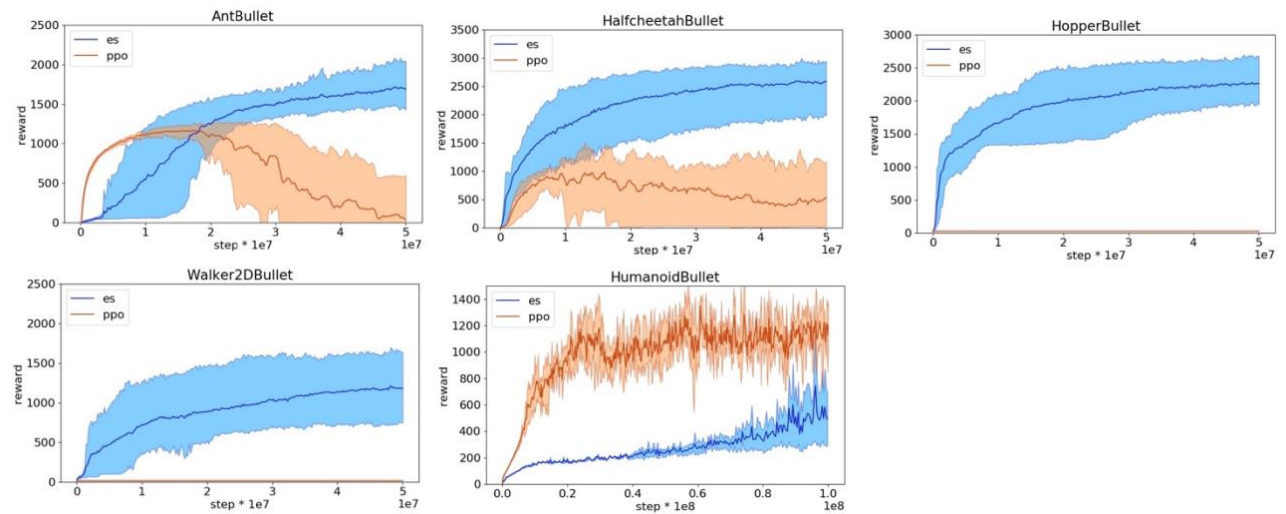

Figure 4. Reward obtained on the Pybullet problems with reward functions optimized for evolutionary strategies. Data obtained with the OpenAI-ES (es) and PPO algorithms. Mean and 90% bootstrapped confidence intervals of the mean (shadow area) across 10 replications per experiment.

We thus attempted to design reward functions suitable for evolutionary strategies. In our first attempt we used a simple rewards function that only rates the agents for their speed toward the destination, i.e. the first component of the reward functions optimized for reinforcement learning. This reward function works well for the HopperBullet, Walker2DBullet, and HalfcheetahBullet problems in which the agents cannot bend laterally (see Figure 4 and 5) but leads to poor performance in the other problems and poor performance in some replications of the HalfcheetahBullet. The problem in the case of the HalfcheetahBullet originates from the fact that in some replications the agents do not use properly all actuated joints. This problem can be solved by including a second component in the reward function of the HalfcheetahBullet, which punishes the agent with -0.1 for every joint that is at its limit (i.e. the component 5 of the reward function optimized for reinforcement learning). With the addition of this second component, the agents evolve a rather effective behavior in all replications (see Figure 4 and 5).

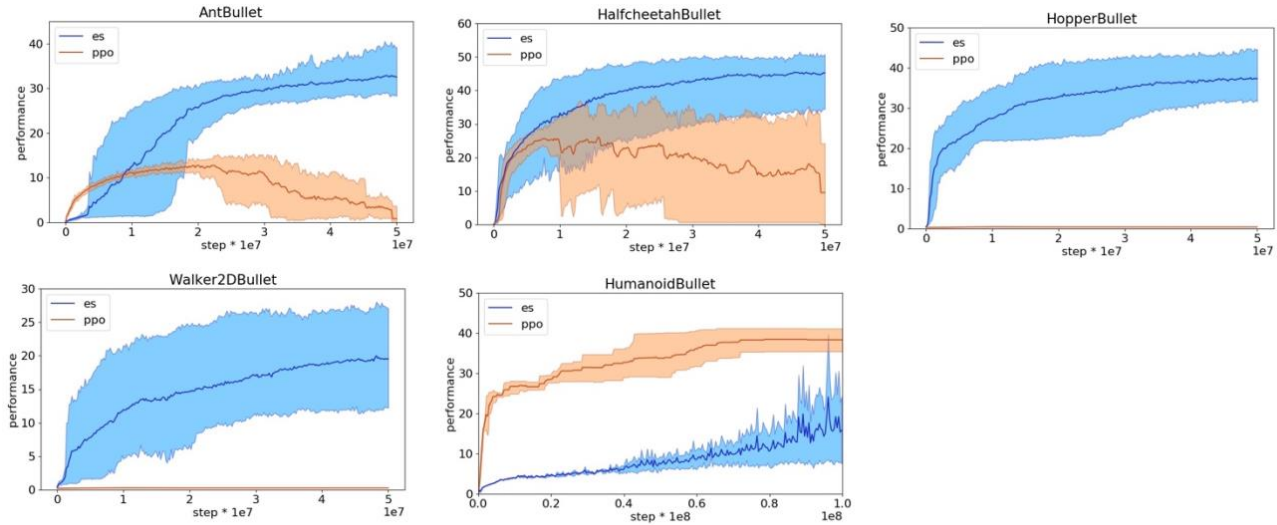

Figure 5. Performance obtained on the Pybullet problems with the reward functions optimized for evolutionary strategies. Data obtained with the OpenAI-ES (es) and PPO algorithms. Performance refers to the distance travelled toward the target destination during an episode in meters. Mean and 90% bootstrapped confidence intervals of the mean (shadow area) across 10 experiments per run.

The reward function with 2 components illustrated above produces effective behaviors in the case of the AntBullet only in some of the replications. In the other replications, instead, the agents slide forward and then fall down or remain still. This problem can be solved by rewarding the robots with a small bonus of 0.01 and by punishing the robots with a small stall cost of -0.01, i.e. by adding two additional components included in the reward functions optimized for reinforcement learning weighted for much smaller constants. This reward function with four components permits to achieve rather good performance in all replications (see Figure 4 and 5).

The reward functions illustrated above are not effective in the case of the HumanoidBullet. The problem, in this case, is that the robots develop physically unrealistic jumping behaviors that enable the robot to start moving fast toward the destination and then to fall off. These unrealistic behaviors are generated by keep pushing the joints over their limits for several steps and by then inverting the direction of movement of the motors so to sum the torque generated by the motors and the torque generated by the simulator to contrast the movement of the joints over the limits. This problem can be solved by adding a fifth component that punishes the robot for pushing the joints over their limits. The development of robust behaviors that reduce the risk of falling down can be further facilitated by increasing the bonus to 0.75. Finally, the development of behaviors that capable of turning left or right in the direction of the target can be facilitated by including a sixth component that consists in the offset of the orientation of the robot with respect to the target in radians weighted for -0.1. This reward function with six components permits to achieve rather good performance in most of the replications (see Figure 4 and 5).

We then tested the PPO with the reward function optimized for the OpenAI-ES method described above. The performance obtained by the PPO with the new reward functions are rather poor and much worse than the performance achieved with the original reward functions (Figure 4 and 5) with the exception of the HumanoidBullet problem. In the case of the HalfcheetahBullet and AntBullet problems the low performance is caused by the fact that the learning process becomes unstable. Indeed the reward (Figure 4) and the distance travelled (Figure 5) decreases during the learning process after an initial phase of improvement. The low performance on the HopperBullet and WalkerBullet problems, instead, reflects the inability to start improving. The usage of a large bonus, in the range of that used in the original reward functions, constitutes an important pre-requisite to start progressing and to avoid a degeneration of the learning process for the PPO. The additional cost components introduced in the reward function optimized for reinforcement learning also help reducing these problems.

The fact that the robots trained with the OpenAI-ES achieve low and high performance with the original and modified reward functions and vice versa, the robots trained with the PPO achieve high and low performance with the original and modified reward functions (with the exception of the HumanoidBullet problem) indicates that reward functions suitable for a reinforcement learning algorithm are not necessarily suitable for an evolutionary strategies and vice versa. Further investigations are necessary to verify whether this is true for all algorithms. However, our data demonstrates that this happens at least in the case of the two state-of-the-art algorithms considered. This implies that a proper comparison of algorithms of different classes should involve the usage of reward functions optimized for each class. Moreover, it implies that comparisons carried out by using reward functions optimized for one method only could be biased.

We hypothesize that this qualitative difference between the OpenAI-ES and the PPO algorithm is caused by the usage of deterministic versus stochastic policies. Evolutionary methods introduce stochastic variations in the policy parameters, across generations, and consequently do not need to use stochastic policies. The state of the actuators is perturbed only slightly through the addition of Gaussian noise with a distribution of 0.01. Reinforcement learning methods, instead, introduce variations through the usage of stochastic policies. The usage of stochastic policies can explain why the agents trained with the PPO do not converge on sub-optimal strategies consisting in standing still in a specific posture without walking despite they receive a significant bonus just for avoiding falling off. Moreover, the usage of stochastic policies can explain why agents trained with PPO have more difficulties to start progressing their ability to walk without being rewarded also for the ability to avoid falling off. Progress in the ability to avoid falling constitute a necessary pre-requisite for developing an ability to walk for agents with stochastic policies. Agents provided with deterministic policies, instead, can develop an ability to walk directly, without first improving their ability to avoid falling off and tend to converge on standing still behavior when rewarded with large bonus for avoid falling off.

The reasons explaining why the OpenAI-ES method outperforms PPO in the HopperBullet, HalfcheetahBullet, AntBullet and WalkerBullet while the PPO method outperforms the OpenAI-ES method on the HumanoidBullet problem deserve further analysis.

## 6 Sensitivity to hyper-parameters

Finally, in this section we analyze the impact of hyper-parameters on the OpenAI-ES method, i.e. the evolutionary method that achieved the best performance.

Table 3 reports the results of two series of ablation experiments carried without the weight-decay normalization (Ng, 2004) and without the virtual batch normalization (Salimans et al., 2016-2017), and four series of experiments carried by varying the size of the population in the range [40, 500] (the total number of evaluation steps is kept constant). The experiments have been carried out on the PyBullet locomotion problems.

Table 3. Average cumulative reward of the best agents of each replication post-evaluated for 3 episodes. Each experiment has been replicated 10 times. Data indicate the average performance and the standard deviation for each experimental condition. The experiments have been continued for $1 * 10_8$ evaluation steps in the case of the HumanoidBullet and for $5 * 10_7$ evaluation steps in the case of the other problems. Data in grey indicate the control conditions that produced a significantly lower performance with respect to the standard condition.

|  | HopperBullet | HalfcheetahBullet | AntBullet | WalkerBullet | HumanoidBullet |
|---|---|---|---|---|---|
| **Standard** | $2293.4 \pm 315.5$ | $2570.7 \pm 460.4$ | $1421.7 \pm 493.2$ | $1232.2 \pm 376.6$ | $1866.9 \pm 307.8$ |
| **No-weight-decay** | $2338.0 \pm 276.3$ | $2396.0 \pm 575.1$ | $46.3 \pm 21.5$ | $1234.9 \pm 577.8$ | $182.2 \pm 19.2$ |
| **No-input-norm.** | $1968.8 \pm 252.5$ | $2224.5 \pm 316.4$ | $1433.7 \pm 328.7$ | $644.1 \pm 311.8$ | $223.1 \pm 17.4$ |
| **Poposize = 40** | $2293.4 \pm 315.5$ | $2570.7 \pm 460.4$ | $1421.7 \pm 493.2$ | $1232.2 \pm 376.6$ | $176.2 \pm 285.7$ |
| **Popsize = 100** | $1953.7 \pm 706.1$ | $2474.5 \pm 464.1$ | $1401.6 \pm 636.3$ | $996.4 \pm 535.5$ | $135.5 \pm 6.1$ |
| **Popsize = 200** | $2044.9 \pm 405.8$ | $2674.0 \pm 265.2$ | $1572.6 \pm 452.4$ | $975.5 \pm 439.4$ | $180.7 \pm 26.4$ |
| **Popsize = 500** | $1773.4 \pm 371.3$ | $2122.8 \pm 1001.5$ | $918.0 \pm 360.5$ | $636.5 \pm 424.0$ | $1866.9 \pm 307.8$ |

As can be seen, the weight decay and virtual batch normalization play an important role in more complex problems, i.e. in the problems in which performance grows more slowly across generations. Indeed, the lack of weight decay leads to significantly lower cumulative reward in the case of the AntBullet and HumanoidBullet problems (Mann-Whitney U test with Bonferroni correction, p-value < 0.05). Moreover, the lack of input normalization leads to significantly lower cumulative reward in the case of the WalkerBullet and HumanoidBullet problems (Mann-Whitney U test with Bonferroni correction, p-value < 0.05). Performance does not significantly differ in the other cases (Mann-Whitney U test with Bonferroni correction, p-value < 0.05).

The analysis of the impact of the population size indicates that the OpenAI-ES method is rather robust with respect to variations of this parameter. Indeed, in the case of the HoppeBullet and HalfcheetahBullet problems, the agents achieved similar performance with all population sizes (Mann-Whitney U test with Bonferroni correction, p-value > 0.05). It the case of the AntBullet and WalkerBullet problems, the agents achieved similar performance with population size in the range [40, 200] (Mann-Whitney U test with Bonferroni correction, p-value > 0.05) and lower performance in experiments in which the population size was set to 500 (Mann-Whitney U test with Bonferroni correction, p-value < 0.05). The fact that small populations evolved for many generations produce similar performance than large populations evolved for fewer generations indicate the presence of a tradeoff between the accuracy of gradient estimation, that increases with the size of the population, and the number of generations necessary to evolve effective behaviors, that decreases with the size of the population. However, the low performance achieved in the HumanoidBullet problem with population smaller than 500 (Mann-Whitney U test with Bonferroni correction, p-value < 0.05) indicates that the minimum size of the population that permits to achieve good performance might depends on the complexity of the problem.

## 7 Conclusions

We analyzed the efficacy of modern neuro-evolutionary strategies for continuous control optimization on the MuJoCo locomotion problems, that constitute a widely used benchmark in the area of evolutionary computation and reinforcement learning, and on additional qualitatively different problems. The term modern evolutionary strategies indicates algorithms that compute the interrelated dependencies among variations of better individuals, or that use a form of finite difference method to estimate the local gradient of the fitness function.

The results obtained on the MuJoCo, Long double-pole and Swarm foraging problems indicate that these methods are generally effective. The comparison of the results obtained with different algorithms indicate that the OpenAI-ES algorithm outperforms or equals the CMA-ES, sNES, and xNES methods on all considered problems.

Overall, the data collected, the ablation studies and the experiments conducted by varying the population size indicate that the efficacy of the OpenAI-ES method is due to the incorporation of optimization and normalization techniques commonly used in neural network research. More specifically, the efficacy of the method can be ascribed to the utilization of the Adam stochastic optimizer, which operates effectively also in the presence sparse gradients and noisy problems and avoids an uncontrolled growth of the size of the connection weights. Moreover, the efficacy of the method can be ascribed to the usage of normalization techniques that preserve the adaptability of the network and reduce overfitting.

Finally, we demonstrate that the reward functions optimized for the PPO are not effective for the OpenAI-ES algorithm and, vice versa, the reward functions optimized for the latter algorithm are not effective for the former algorithm. This implies that the reward function optimized for a reinforcement learning algorithm are not necessarily suitable for an evolutionary strategy algorithm and vice versa. Consequently, this implies that a proper comparison of algorithms of different classes should involve the usage of reward functions optimized for each algorithm. Indeed, comparisons carried out by using reward functions optimized for one method only could be biased. The usage of deterministic policies (commonly used in evolutionary methods) versus stochastic policies (commonly used in reinforcement learning methods) seems to be an important cause of the differences observed between the OpenAI-ES and the PPO algorithms with respect to the sensitivity to the reward function. Whether the usage of different reward functions is necessary for all evolutionary and reinforcement learning algorithms or only for some algorithms deserve further investigations. Similarly, the identification of the characteristics that make a reward function suitable for a specific algorithm deserves further investigations.

**Appendix**

The parameters used in the experiments are summarized in Table 4. The network policy is constituted by a feed-forward neural network in all cases with the exception of the Long double-pole and the Swarm foraging problems that requires memory. The number of internal layers is set to 1 in all cases with the exception of the Humanoid and BulletHumanoid problems that can benefit from the usage of multiple layers. The activation function for the internal neurons is the hyperbolic tangent function (tanh) in all cases. The activation function of the output neurons is the linear function in the case of the locomotor problems and the tanh function in the case of the Long double-pole and Swarm foraging problems. The linear function generally works better, since it produces more varied response between the individuals of the population, but it does not normalize the output in a limited range. For this reason the tanh function has been used in the Long double-pole and Swarm foraging problems which require a bounded range. Following Salimans et al. (2017), we use a binary encoding in which 10 output neurons are used to select one of 10 different activation values for each output in the case of the Swimmer and the Hopper. This did not turned to be necessary in the case of the BulletHopper, probably due to the optimization of the reward functions.

Table 4. Parameters used in all experiments.

|  | architecture | internal layers | activation function | output activation function | normalization | action noise | evaluation episodes | popsize | total steps |
|---|---|---|---|---|---|---|---|---|---|
| Swimmer | Feed-forward | 1 | tanh | 10 bins | Yes | 0.01 | 1 | 40 | $5 * 10^7$ |
| Hopper | Feed-forward | 1 | tanh | 10 bins | Yes | 0.01 | 1 | 40 | $5 * 10^7$ |
| Halfcheetah | Feed-forward | 1 | tanh | linear | Yes | 0.01 | 1 | 40 | $5 * 10^7$ |
| Walker2D | Feed-forward | 1 | tanh | linear | Yes | 0.01 | 1 | 40 | $5 * 10^7$ |
| Humanoid | Feed-forward | 2 | tanh | linear | Yes | 0.01 | 1 | 500 | $2.5 * 10^8$ |
| Long double-pole | Recurrent | 1 | tanh | tanh | No | 0.0 | 50 | 40 | $1 * 10^{10}$ |
| Swarm foraging | Recurrent | 1 | tanh | tanh | No | 0.0 | 6 | 40 | $1.5 * 10^6$ |
| BulletHopper | Feed-forward | 1 | tanh | linear | Yes | 0.01 | 1 | 40 | $5 * 10^7$ |
| BulletHalfcheetah | Feed-forward | 1 | tanh | linear | Yes | 0.01 | 1 | 40 | $5 * 10^7$ |
| BulletAnt | Feed-forward | 1 | tanh | linear | Yes | 0.01 | 1 | 40 | $5 * 10^7$ |
| BulletWalker2d | Feed-forward | 1 | tanh | linear | Yes | 0.01 | 1 | 40 | $5 * 10^7$ |
| BulletHumanoid | Feed-forward | 2 | tanh | linear | Yes | 0.01 | 3 | 500 | $10 * 10^8$ |

The virtual batch normalization was used in all the locomotors environments. This since the range of the velocity of the joints encoded in some of the sensors vary widely during the evolutionary process. The virtual batch normalization was not used in the other problems that are less affected by this issue. Noise was added to the motors in the case of the locomotors problems, but not in the case of the other problems. The initial conditions of the agents in the Long double pole and Swarm foraging problems is already highly varied. Consequently, evolution tends to produce solutions that are robust to variations independently of the addition of noise. The number of evaluation episodes is set to 1 in all locomotors problems and to 6 and 50 in the case of the Swarm Foraging and Long Double Pole problems which expose agents to higher variations of the environmental conditions. The number of episodes has been set to 3 in the case of the HumanoidBullet problem since small variations in the initial posture tend to have a significant impact on the stability of these robots.

The size of the population is set to 40 in all cases with the exception of the Humanoid and HumanoidBullet. This parameter has been varied systematically in the case of the BulletLocomotors (see Table 3). Using larger populations generally produces similar results in these problems but it seems to be a necessity in the case of the Humanoid problem. The duration of the evaluation process has been set on the basis of the complexity of the problem and on the time cost of the simulation. For example, the experiments in the case of the Long double pole problems could be continued for a rather high number of total evaluations ($1 * 10^{10}$) because performance keep increasing significantly in the long term and because this is feasible in terms of simulation time. In general we continued the evolutionary process for significantly longer periods of time with respect to previous published works to ensure that the difference in performance does not reflect only the evolution speed but also the quality of the solutions obtained in the long term.

For the experiment performed with the PPO algorithm we used the default parameters included in baseline (https://github.com/openai/baselines).